%% file: iclr2016_conference.tex
\documentclass{article} 
\usepackage{iclr2016_conference,times}
\usepackage[hidelinks]{hyperref}
\usepackage{url}

\usepackage{graphicx} 
\usepackage{subfigure} 

\usepackage{amsmath}

\title{Dropout as data augmentation}

\author{
Xavier Bouthillier \thanks{Both authors contributed equally}\\
Universit\'{e} de Montr\'{e}al, Canada \\
\texttt{xavier.bouthillier@umontreal.ca} \\
\And
Kishore Konda $^*$\\
Goethe University Frankfurt, Germany\\
\texttt{konda.kishorereddy@gmail.com} \\
\AND
Pascal Vincent \\
Universit\'{e} de Montr\'{e}al, Canada and CIFAR\\
\texttt{pascal.vincent@umontreal.ca} \\
\And
Roland Memisevic \\
Universit\'{e} de Montr\'{e}al, Canada \\
\texttt{roland.memisevic@umontreal.ca} \\
}

\begin{document}

\maketitle

\begin{abstract}
Dropout is typically interpreted as bagging a large number of models sharing
parameters. We show that using dropout in a network can also be interpreted as 
a kind of data augmentation in the input space without domain knowledge.
We present an approach to projecting the dropout noise within a network back
into the input space, thereby generating augmented versions of the training data,
and we show that training a deterministic network on the augmented
samples yields similar results. Finally, we propose a new dropout noise 
scheme based on our observations and show that it improves dropout results 
without adding significant computational cost. 
\end{abstract}

\section{Introduction}
\label{sec:intro}
\input{./text/intro.tex}

\section{Dropout}
\label{sec:noise}
\input{./text/noise.tex}

\section{From a data augmentation perspective}
\label{sec:dataaug}
\input{./text/dataaug.tex}

\section{Improving the set of augmentations}
\label{sec:improve}
\input{./text/improve.tex}

\section{Experiments}
\label{sec:experiments}
\input{./text/experiments.tex}

\section{Related Work}
\label{sec:discussion}
\input{./text/related.tex}

\section{Conclusion}
\label{sec:conclusion}
\input{./text/conclusion.tex}

 \section*{Acknowledgements}
This work was supported by the German Federal Ministry of Education and
 Research (BMBF) in the project 01GQ0841 (BFNT Frankfurt), an NSERC 
Discovery grant, a Google faculty research award and Ubisoft. 
We are grateful towards the developers of Theano 
\citep{bergstra2010theano,bastien2012theano}, Fuel and Blocks 
\citep{van2015blocks}.

\bibliography{literature}
\bibliographystyle{iclr2016_conference}

\newpage
 \section{Appendix}
 \label{sec:supp}
 \input{./text/appendix.tex}

\end{document}

%% file: text/intro.tex
Noise is normally seen as intrinsically undesirable. The word itself bears a 
very negative connotation. It is not surprising then that many early mathematical
models in neuroscience aimed to factor out noise by any means. A few decades ago, 
the use of stochastic resonance \citep{stochasticresonancebiology} in neuro-scientific 
models initiated a new interest in neurosience regarding random fluctuations and
the role they play in the brain. Theories about neuronal noise are now 
flourishing and previous deterministic models are improved by the 
incorporation of noise \citep{signalornoise}.

Biological brains have always been a strong inspiration when it comes to developing
learning algorithms. Considering the amount of noise which takes place in the brain
during learning, one can wonder if this has any beneficial effect. Many techniques
in machine learning have made use of noise to improve performance recently, namely, 
Denoising Autoencoders \citep{dae}, dropout \citep{dropout}
and its relative, DropConnect \citep{dropconnect}. Those successful approaches suggest that
neuronal noise plays a fundamental role in the process of learning and should be
studied more thoroughly. 

Using dropout can be viewed as training a huge number of neural networks with 
shared parameters and applying bagging at test time for better 
generalization \citep{baldi2013understanding}. 
Binary noise can also be viewed as preventing neurons from co-adapting, which improves 
the generalization of the model even more. 
In this paper, we propose an alternative view and suggest noise schemes like 
dropout are implicitly incorporating a form of sophisticated data augmentation.
In Section \ref{sec:dataaug}, we formulate a method to generate data which 
replicates dropout noise within a deterministic network, and demonstrate in 
Section \ref{sec:experiments} that there is no significant loss
of accuracy.

Finally, capitalizing on the idea of data augmentation, we present in section
\ref{sec:improve} an extension of dropout which uses random noise levels to 
improve the variety of samples. 
This simple extension improves classification performance across different 
network architectures, yielding competitive results on the MNIST permutation 
invariant classification task.

%% file: text/noise.tex
The main goal when using dropout is to regularize the neural network we are 
training. The technique consists of dropping neurons randomly with some
probability $p$. Those random modifications of the network's stucture
are believed to avoid co-adaptation of neurons by making it impossible
for two subsequent neurons to rely solely on each other 
\citep{srivastava2014dropout}. The most accepted interpretation of dropout 
is that is implicitely bagging at test time a large number of neural networks 
which share parameters.

Assume $h(x)$ is a linear projection of a $d_i$-dimensional input $x$ into a 
$d_h$-dimensional space:
\begin{equation}
    h(x) = xW + b
\end{equation}

Given $a(h)$ and $\tilde{a}(h)$, an activation function and its
noisy version where $M \sim \mathcal{B}(p_{h})$ and $\mbox{rect}(h)$ is a rectifier 

\begin{align} 
    \label{eq:dropout}
    a(h) &= \mbox{rect}(h) \\
    \tilde{a}(h) &= M \odot \mbox{rect}\left(h\right) \label{eq:tilde_a}
\end{align}

Eq. \ref{eq:tilde_a} denotes the activation with dropout during training and
eq. \ref{eq:dropout} the equation of the activation at test time.
\cite{srivastava2014dropout} suggest to scale the activations $a(h)$ with $p$
at test time to get an approximate average of the unit activation.

%% file: text/dataaug.tex
In many previous works it has been shown that augmenting data by using 
domain specific transformations helps in learning better 
models \citep{lecun1998gradient, simard2003best, krizhevsky2012imagenet, ciresan2012multi}.
In this work, we analyze dropout in the context of data augmentation. 
Considering the task of classification,
given a set of training samples, the objective would be to learn a mapping 
function which maps every input to its corresponding output label. To generalize, the mapping 
function needs to be able to correctly map not just the training samples 
but also any other samples drawn from the data distribution.  
This means that it must not only map input space sub-regions represented by 
training samples, but all high-probability sub-regions of the natural distribution.

One way to learn such a mapping function is by augmenting the training data such that
it covers a larger portion of the natural distribution. Domain-based data augmentation
helps to artificially boost training data coverage 
which makes it possible to train a better mapping function.
We hypothesize that noise based regularization techniques result in a
similar effect of increasing training data coverage at every hidden layer and 
this work presents multiple experimental observations to support our hypothesis.

\subsection{Projecting noise back into the input space}
\label{subsec:noise_projection}

We assume that for a given $\tilde{a}(h)$, there exist an $x^*$, 
such that 

\begin{equation}
    \label{eq:x_star_approx}
    (a \circ h)(x^*) = \mbox{rect}(h(x^*)) \approx \vec{m}\odot \mbox{rect}\left(h(x)\right) = (\tilde{a} \circ h)(x)
\end{equation}

Similarly to adversarial examples from \cite{goodfellow2014explaining}, an $x^*$
can be found by minimizing the squared error $L$ using stochastic gradient descent

\begin{equation}
 \label{eq:x_star_cost}
   L\left(x, x^*)\right) = \left|(a \circ h)(x^*) - (\tilde{a} \circ h)(x)\right|^2
\end{equation}

Equation \ref{eq:x_star_cost} can be generalized to a network with $n$ hidden 
layers. To lighten notation we first define

\begin{align}
    \tilde{f}^{(i)}(x) &= \left(\tilde{a}^{(i)} \circ h^{(i)} \circ \dots \circ \tilde{a}^{(1)} \circ h^{(1)}\right)(x) \label{eq:f_tilde_x}\\
    f^{(i)}(x^*)       &= \left(       a ^{(i)}                 \circ h^{(i)} \circ \dots \circ        a ^{(1)}                 \circ h^{(1)}\right)(x^*) \label{eq:f_x_star}
\end{align}

We can now compute the back projection corresponding to all hidden layer
activations at once, which results in minizing the loss $L$

\begin{equation}
 \label{eq:x_star_cost_n}
 L\left(x, x^{(1)^*}, \dots, x^{(n)^*}\right) = 
 \sum_{i=1}^n \lambda_i \left|f^{(i)}\left(x^{(i)^*}\right) - \tilde{f}^{(i)}(x)\right|^2
\end{equation}

We can show by contradiction that one is unlikely to find a single 
${x^* = x^{(1)^*} = x^{(2)^*} = \dots = x^{(n)^*}}$ that significantly reduces
$L$. The proof is detailed in appendix subsection \ref{subsec:x_star_proof}. 
Fortunately, it is easy to find a different $x^*$ 
for each hidden layer, by providing multiple inputs 
$(x, x^{(1)^*}, x^{(2)^*}, \dots, x^{(n)^*})$, where $n$ is the number of 
hidden layers. As each $x^{(i)^*}$ is the back projection of a transformation 
in the representation space defined by the $i$-th hidden layer, it suggests
viewing dropout as a sophisticated data augmentation procedure 
that samples data around training examples with respect to 
different level of representations.
This raises the question whether we can train the network deterministically on the $x^{(i)^*}$ rather than using dropout. The answer is not trivial, because
\begin{enumerate}
    \item When using $(x, x^{(1)^*}, x^{(2)^*}, \dots, x^{(n)^*})$ as inputs,
        dropout is not effectively applied to every layer at the same time.
        The local stochasticity preventing co-adaptation is then present at a
        specific layer only once for every $x^{(i)^*}$. This could be not 
        aggressive enough to avoid co-adaptation.
    \item The gradients of the linear projections will differ greatly. 
In the case of dropout, $\frac{\partial{h}}{\partial{W^{(i)}}}$ is always equal to its 
input transformation, i.e. $\tilde{f}^{(i-1)}(x)$, whereas the deterministic version of the training 
will update $W^{(i)}$ according to $\left(f^{(i-1)}(x^{(1)^*}), \dots, f^{(i-1)}(x^{(n)^*})\right)$\footnote{Because we train on $n$ samples from $x$, one for each hidden layer}.
\end{enumerate}


Although we proved a single $x^*$ minimising \ref{eq:x_star_cost_n} is 
difficult to find for a large network, 
we show experimentally in section \ref{sec:experiments} that it is possible to do so within reasonable 
approximation for a relatively small two hidden layer network. We further show that dropout can be replicated by projecting
the noise back on the input space without a significant loss of accuracy.

%% file: text/improve.tex
When dealing with domain-based transformations, we intuitively look for the
richest set of transformations. In computer vision for instance, translations, rotations,
scalings, shearings and elastic transformations are often combined. 
Looking at dropout from a data augmentation perspective, this
intuition raises the following question: given that noise scheme used is
implicitely applying some transformations in the input space, which one would produce the
richest set of transformations?

With noise schemes like dropout, there are two important components which
influence the transformations; The probability distribution of $\vec{m}$ and 
the features of the neural network used to encode $h(x)$. Modifying the
probability distribution is the most straighforward way to improve the set of 
transformations and will be the main focus of this paper. However, features
of the neural network play a key role in the transformations and we will
outline some possible avenues in the conclusion section.

\subsection{Random noise levels}

\label{subsec:randomnoise}

While using dropout, the proportion of neurons dropped is very close to
probability $p$. It follows naturally from Binomial distribution's expectation. The
transformations induced are as different as the values $M$ can take.
Despite this, their magnitude is as constant as the proportion of neurons
dropped. That means, every transformations displaces the sample to a
relatively constant distance but in random directions in a high dimensional
space. 

A simple way to vary the transformation magnitude randomly is to replace $p$ by
a random variable. Let $\rho_h \sim \mathcal{U}(0, p_h)$ and 
$M_{hij} \sim \mathcal{B}(\rho_h)$ where $h$ defines the layer, 
$i$ the sample, and $j$ the layer's neuron. It is important to use the
same $\rho$ for all neurons of a layer, otherwise we would have 
$M_{hij} \sim \mathcal{B}(\frac{p_h}{2})$.

To compensate for the change of level of activations during test, a scaling is 
normally applied. One could also simply apply the inverse scaling during
training, turning equation \ref{eq:tilde_a} into

\begin{equation} 
    \label{eq:scaled_tilde_a}
    \tilde{a}(h) = \frac{1}{1-p}M \odot \mbox{rect}\left(h\right)
\end{equation}

To adapt the equation to random dropout level, we simply need to replace $p$ with $\rho$

\begin{equation} 
    \label{eq:scaled_random_tilde_a}
    \tilde{a}(h) = \frac{1}{1-\rho}M \odot \mbox{rect}\left(h\right)
\end{equation}

No scaling needs to be done during test anymore.

\begin{figure}
    \center
        {\includegraphics[width=0.45\textwidth]{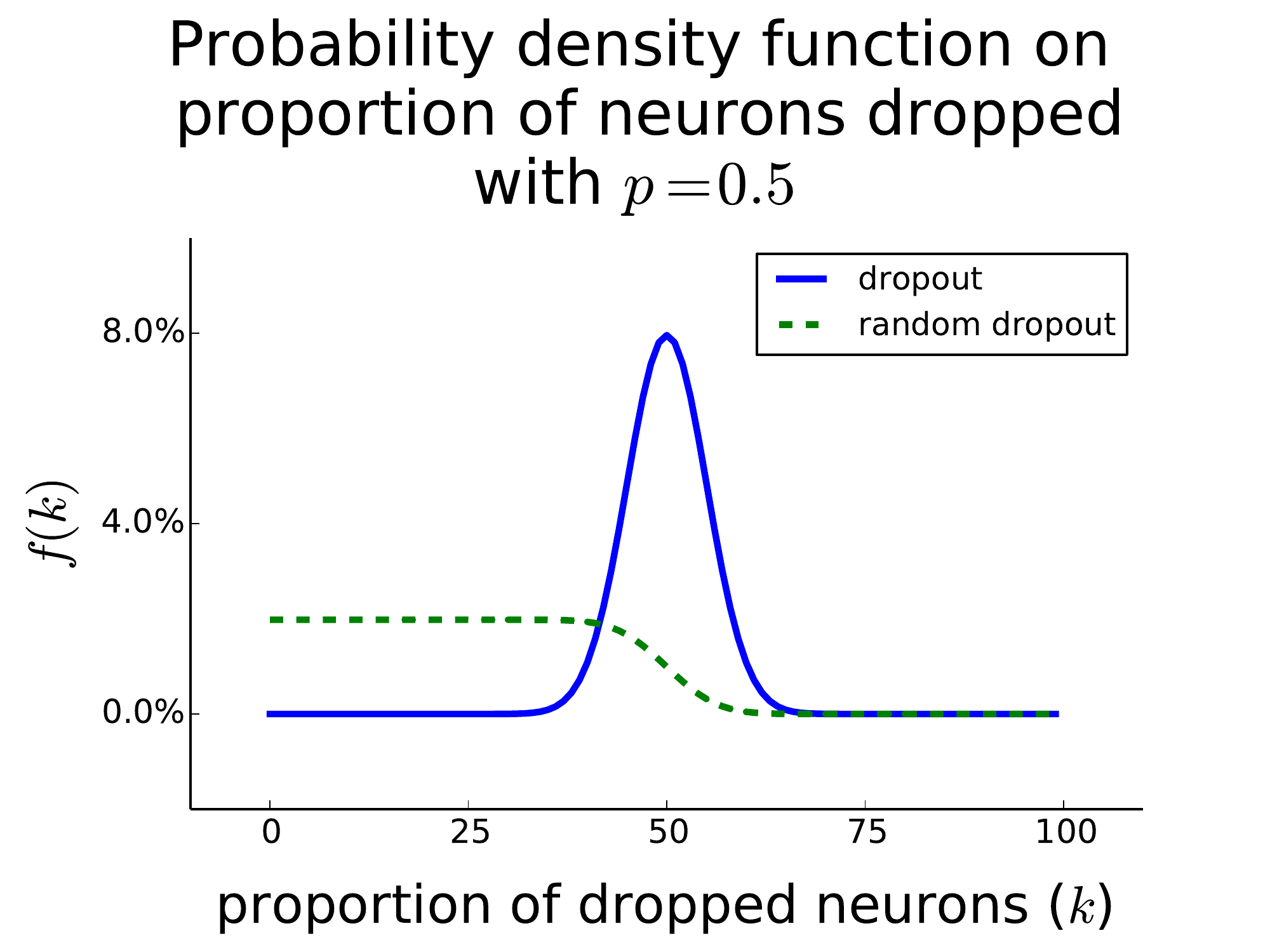}
    }
    \caption{The density function on proportion of neurons dropped is very peaked
             around $p$ for dropout. This results in a low variance of the
             transformations magnitude induced by dropout. Random dropout on
             the other hand has a constant density function under $pN$, $N$
             being the number of neurons, because
             $\rho \sim \mathcal{U}(0, p)$. It is thus more likely to see
             transformations closer to identity while it is very unlikely for
             standard dropout. This is intuitively desirable, as much as varying 
             translation distances is preferable over constant large distances.}
    \label{fig:noise_density_function}
\end{figure}

Figure \ref{fig:noise_density_function} shows the differences between density
function on proportion of neurons dropped for dropout and random dropout.
Transformations induced by random dropout are clearly more diverse than those
induced by dropout.

%% file: text/experiments.tex
\subsection{Visualizations of noise projected back into the input space}
\label{subsec:exp_visu}
\input{./text/visualizations.tex}

\subsection{Equivalence of dropout and noisy samples}
\label{subsec:exp_aug_data}
\input{./text/train.tex}

\subsection{Richer noise schemes}
\label{subsec:exp_random_noise}
\input{./text/random_dropout.tex}

%% file: text/visualizations.tex
Visualizing the noise projected back into the input space helps to understand
what kind of transformations are induced by dropout. Unsupervised models
learn more general features than supervised fully connected neural networks
and produce thus more visually appealing transformations. For this reason, we
trained autoencoders with dropout on the hidden layer to generate samples of
transformations.

The autoencoder is very similar to denoising autoencoders, the only difference
is that a Bernoulli mask is applied to the hidden activations rather than 
to the input. There is thus no noise applied to the input explicitly. Models
are trained for 300 epochs, with mini-batch size of 100, $p$ = 0.4, 
a learning rate of 0.001 on MNIST,
0.0001 on CIFAR-10 and a momentum of 0.7 on MNIST, 0.5 on CIFAR-10. For
CIFAR-10, we do preprocessing with PCA dimensionality reduction and retain 781 
features.

\begin{figure}
    \center
    \subfigure[MNIST]
        {\includegraphics[width=0.40\textwidth]{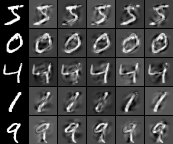}
    }
    \hspace{2em}
    \subfigure[CIFAR]
        {\includegraphics[width=0.40\textwidth]{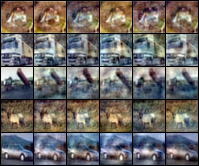}
    }
    \caption{Visualization of noisy samples on MNIST and CIFAR-10.
             The first column represent original samples from MNIST and CIFAR-10
             datasets. Each row contains samples from the same original sample.
             Every other column represents noisy samples produced by
             back-projecting the noise into the input space. Each one 
             is induced by a different noise mask, i.e. a different value of
             $M$.}
    \label{fig:x_star_samples}
\end{figure}

Once the model is trained, we use gradient descent to compute $x^*$ as 
described in \ref{subsec:noise_projection}. We iterate for 10 epochs with
a learning rate of 100 for both MNIST and CIFAR-10. Figure \ref{fig:x_star_samples}
shows well how close $x^*$ are from the natural input space and we 
clearly see that the classes are still distinguishable.

\begin{figure}
    \center
    \subfigure[]
        {\includegraphics[width=0.20\textwidth]{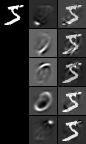}
    }
    ~
    \subfigure[]
        {\includegraphics[width=0.20\textwidth]{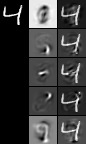}
    }
    ~
    \subfigure[]
        {\includegraphics[width=0.20\textwidth]{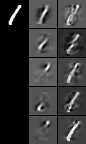}
    }
    ~
    \subfigure[]
        {\includegraphics[width=0.20\textwidth]{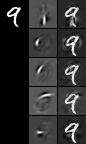}
    }
    \caption{Visualization of the influence of the features on the transformations
             induced by dropout on MNIST. First column represent original samples 
             from MNIST dataset. Second column represents the five most active
             features on each given original sample. Last column represents 
             noisy samples produced by back-projecting the noise into the input
             space while keeping the selected feature shut off to 0. We can see
             that the features are not simply removed from the input, but
             rather {\it destroyed} in such a way that other features 
             highly dependant on the same subregion still have the same 
             activation level.}
    
    \label{fig:feature_samples}
\end{figure}

To help understand how each feature influences the transformation, we isolate
five most active features for a given input and shut them off, each separately.
For each feature shut off, we compute $x^*$ using gradient descent. Figure
\ref{fig:feature_samples} shows the results found for MNIST. One
could think the features dropped by the noise are simply removed from the
input. It turns out removing the feature would affect the activation of other
neurons. Because of this, features are rather {\it destroyed} in the input in
such a way that other features highly dependant on the same subregion still 
have the same activation level.

%% file: text/train.tex
We ran a series of experiments with fully connected feed forward neural 
networks on the MNIST and CIFAR-10 datasets to study the effect of replacing 
dropout by corresponding noisy inputs.
Each network consists of two hidden layers with rectified linear units 
followed by a {\it softmax} layer. We experimented with four different
architectures each one with a different number of units in the hidden layers: 
2500-2500, 2500-1250, 1500-1500 and 1500-750. 

The MNIST dataset consists of $60000$ training samples and $10000$ test 
samples. We split the training set into a set of $50000$ samples for training
and $10000$ for validation. Each network is trained for $501$ epochs and the 
best model based on validation error is selected. Later the best model is 
further trained on the complete training set of $60000$ samples 
(training + validation split) for another $150$ epochs. The mean error 
percentage for the last $75$ epochs is used for comparison. 

We also ran experiments on the CIFAR-10 permutation invariant task using the
same network architectures described above. The dataset consists of $50000$ 
training samples and $10000$ test samples. We use PCA based dimensionality 
reduction without whitening as preprocessing scheme, retaining 781 features.
We used the same approach as in the MNIST experiments to train the 
networks and for reporting the performances.

At each epoch, an $x^*$ is generated for each training sample. It proved to be possible to find 
good $x^*$ approximations for the entire network at once for a 2-hidden layer network. 
Thus, we trained on $x$ and $x^*$ solely rather than $x$, $x^{(1)^*}$ and 
$x^{(2)^*}$ as it gave a significant speed up. 
For simplicity, the network is trained on $x$ for an epoch than on $x^*$ for an epoch. All $x^*$ 
are generated with parameter values of the model at the beginning of the epoch.

\begin{figure}
 \center
 \subfigure[]{\includegraphics[width=0.33\textwidth]{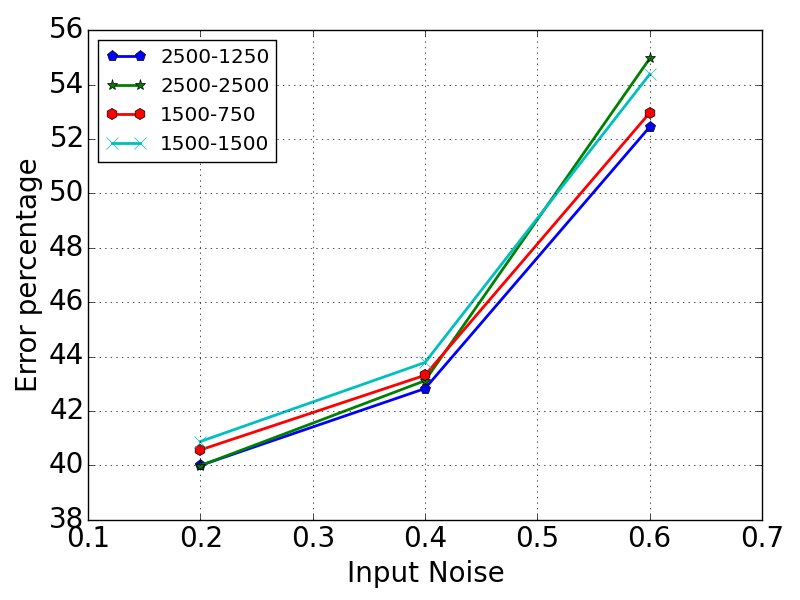}}
 \subfigure[]{\includegraphics[width=0.33\textwidth]{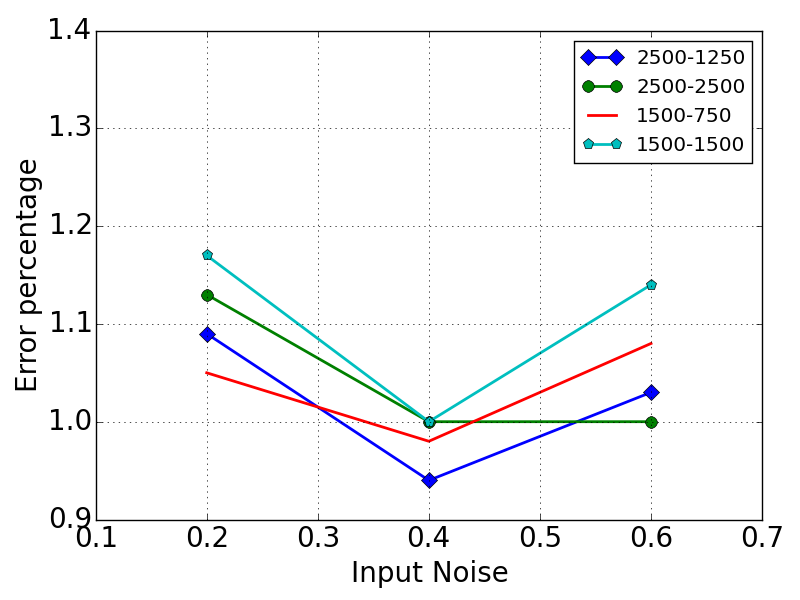}}
 
 \subfigure[]{\includegraphics[width=0.33\textwidth]{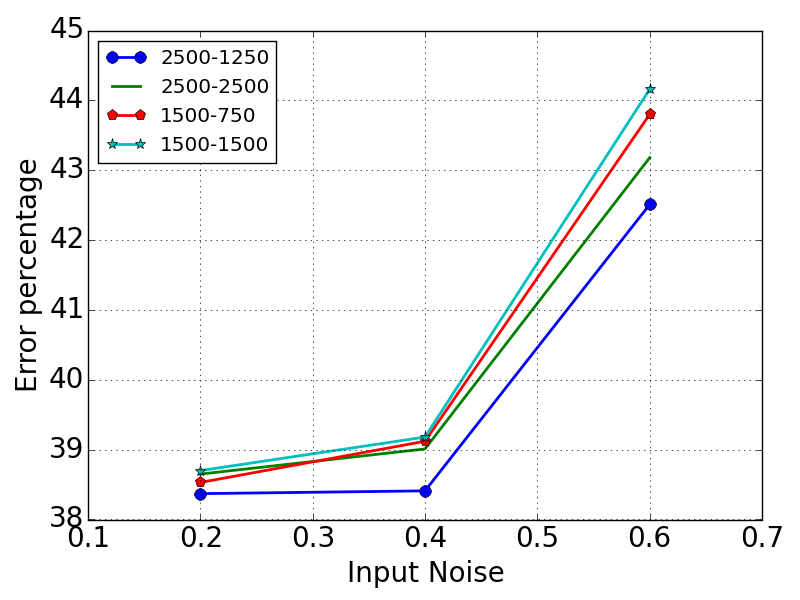}}
 \subfigure[]{\includegraphics[width=0.33\textwidth]{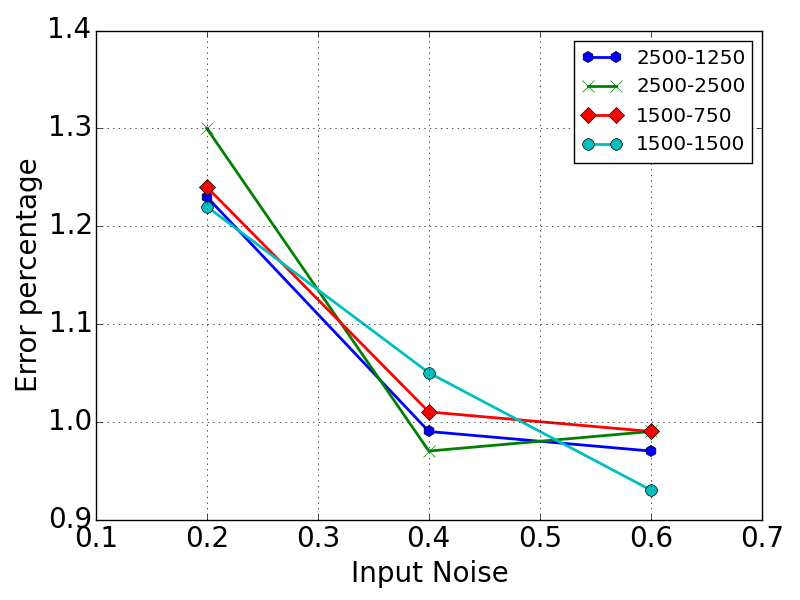}}
 \caption{Error percentages on the MNIST and CIFAR-10 datasets using MLP architectures trained with different corruption schemes.
 {\bf Row-1}: Experiments using dropout. {\bf Row-2}: Experiments using noise back projection. {\bf Column-1}: CIFAR-10
 {\bf Column-2}: MNIST. }
 \label{fig:pnoise}
\end{figure}

Noisy inputs $x^*$ are found using stochastic gradient descent. 20 learning 
steps are done with a learning rate of 300.0 for first hidden layer
and 30 for second hidden layer. The results for these experiments are shown in figure \ref{fig:pnoise}.

Results suggest that Dropout can be replicated
by projecting the noise back into the input and training 
a neural network deterministically on this generated data. There is not significant drop in 
accuracy, it is even slightly better than Dropout in the case of CIFAR-10.
This supports the idea that dropout can be seen as data augmentation.

%% file: text/random_dropout.tex
We ran a series of experiments with fully connected feed forward neural 
networks on the MNIST and CIFAR-10 datasets to compare dropout.
Each networks consist of two hidden layers with rectified linear units 
followed by a {\it softmax} layer. We experimented with three different network
architectures each one with a different number of units in the hidden layers: 
2500-625, 2500-1250 and 2500-2500.
Each network is trained and validated the same way as mentioned in previous section.

\begin{figure}
 \subfigure[]{\includegraphics[width=0.33\textwidth]{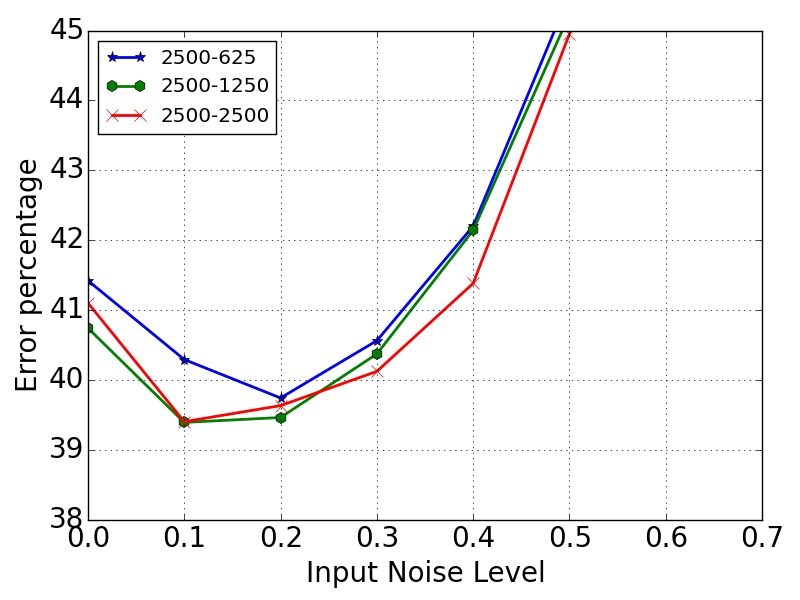}}
 \subfigure[]{\includegraphics[width=0.33\textwidth]{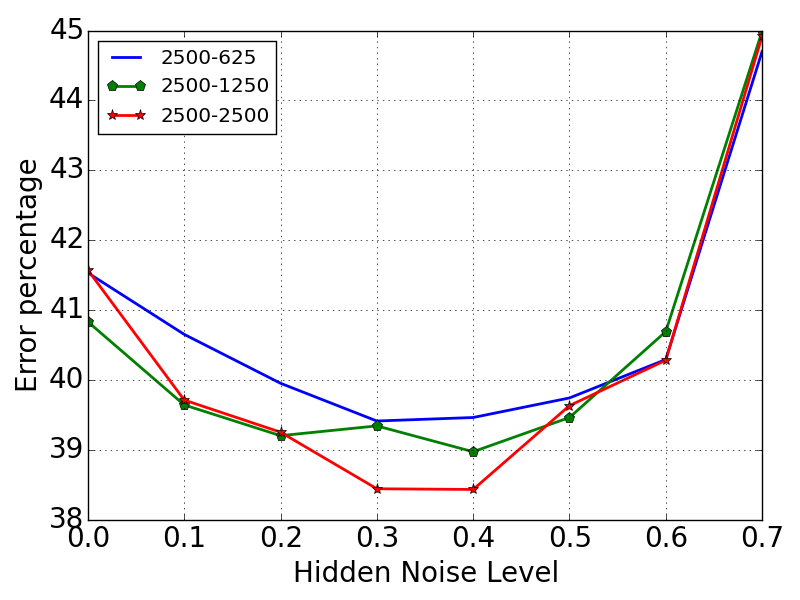}}
 \subfigure[]{\includegraphics[width=0.33\textwidth]{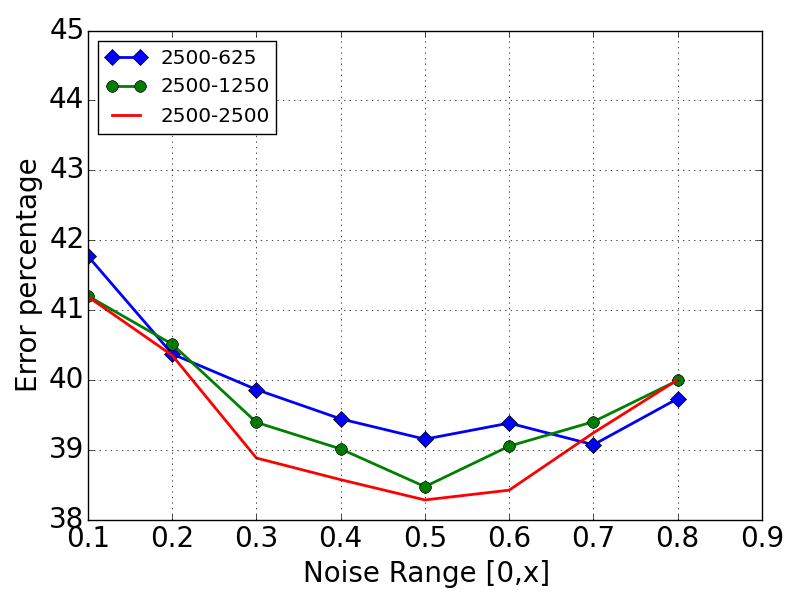}} \\
 
 \subfigure[]{\includegraphics[width=0.33\textwidth]{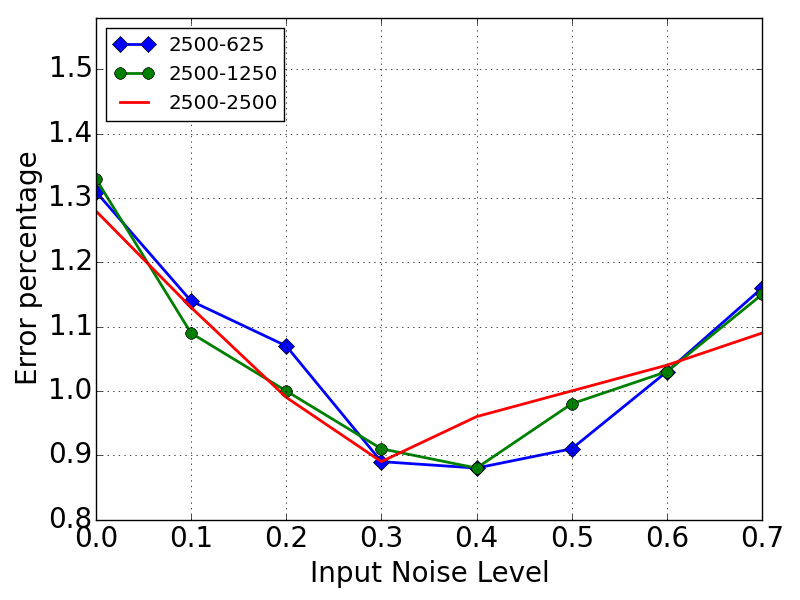}}
 \subfigure[]{\includegraphics[width=0.33\textwidth]{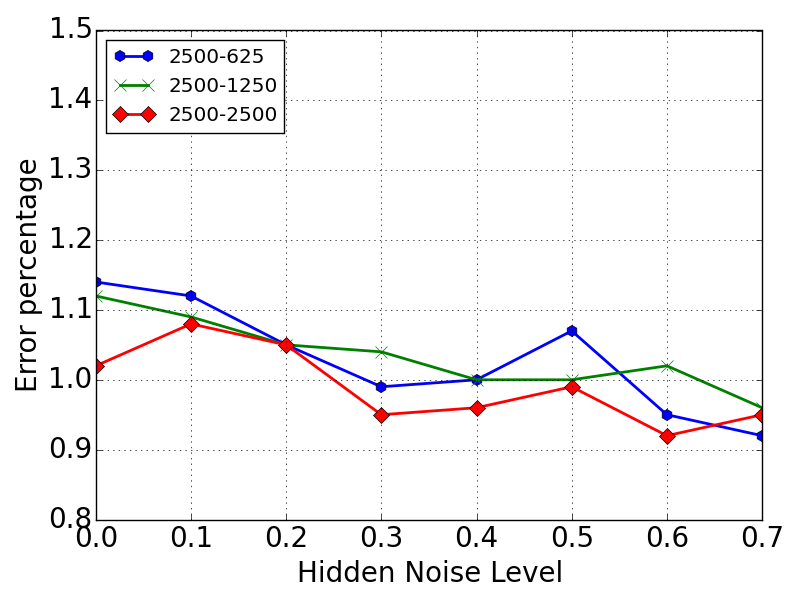}}
 \subfigure[]{\includegraphics[width=0.33\textwidth]{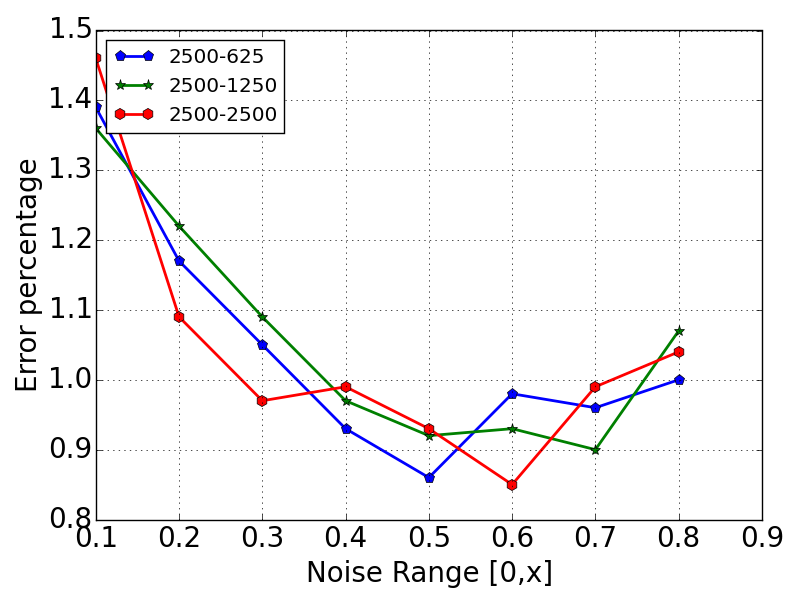}}
 \caption{Error percentages on the MNIST and CIFAR-10 datasets using MLP architectures trained with different corruption schemes.
 {\bf Row-1}: Experiments on CIFAR-10. {\bf Row-2}: Experiments on MNIST. {\bf Column-1}: Using dropout with varying input noise and fixed hidden noise of 0.5. 
 {\bf Column-2}: Using dropout with varying hidden noise with fixed input noise of 0.2. {\bf Column-3}: Using Random-dropout with varying noise range $[0,x]$ used at hidden and input layers.}
 \label{fig:dropout_noise}
\end{figure}

First, we evaluated the dropout noise scheme by training the networks with a fixed 
hidden noise level of $0.5$ and the input noise level varying from $0.0$ to 
$0.7$ with increments of $0.1$ for each experiment.
In the second experiment, we fixed the input noise level at $0.2$ and the hidden 
noise level is varied from $0.0$ to $0.7$, again with an increment of $0.1$. 
In the final set of experiments we use the random dropout noise scheme using 
the same noise level at input and hidden layers.
The noise level in this case is a range $[0,x]$ where $x$ is varied from 
$0.0$ to $0.8$ with increment $0.1$.
The classification performances corresponding to the all the experiments on
both the datasets are reported in Figure \ref{fig:dropout_noise}.

Random dropout improves the performance of the models over dropout with 
no additional computational cost.

\begin{figure}
    \center
    \subfigure[MNIST]
        {\includegraphics[width=0.40\textwidth]{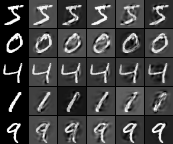}
    }
    \hspace{2em}
    \subfigure[CIFAR]
        {\includegraphics[width=0.40\textwidth]{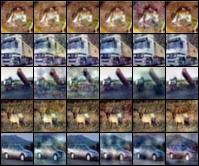}
    }
    \caption{Visualization of noisy samples from random dropout on MNIST and CIFAR-10.
             The first column represent original samples from MNIST and CIFAR-10
             datasets. Each row contains samples from the same original sample.
             Every other column represents noisy samples produced by
             back-projecting the noise into the input space. Each one 
             is induced by a different noise mask and a different noise level, 
             i.e. a different value of $\rho$ and $M$. The transformations from
             figure \ref{fig:x_star_samples} and this one are clearly 
             different. Random dropout applies transformations with different
             strenghs, i.e., the transformed input can be very close to very far
             from the original input while standard dropout
             always applies transformations with the same strengh.
         }
    \label{fig:x_star_samples}
\end{figure}

%% file: text/related.tex
To the best of our knowledge there is no work analyzing dropout from a data
augmentation perspective. Nonetheless, there is a plethora of excellent works
about dropout, some describing his regularisation properties and others developing 
new kind of noise schemes based on different intuitions.

Regularization properties of noise have been known for more than a decade, 
\cite{bishop1995training} showed for instance that the regularization term induced
by noise belongs to the class of generalized Tikhonov regularizers for sum-of-squares 
error functions. More recently, \cite{baldi2013understanding} proved that dropout noise
scheme specifically applies a regularisation term very similar to usual weight decay.

Not exactly about regularization, but uncertainty, \cite{gal2015dropout} gave 
a very inspiring interpretation of dropout as a Bayesian approximation.

Different noise schemes are used by \cite{poole2014analyzing} and applied at
different positions in autoencoders; input, pre-activation and 
post-activations. They report better results than denoising autoencoders and 
also support that Gaussian noise yields better results than dropout on MNIST 
classification task.



\cite{bachman2014learning} emphasise bagging interpretation of dropout
and propose a generalization called pseudo-ensembles and a related regularizer
which makes it possible to train semi-supervised networks.

Some recent work reports that noise level schedules, inspired by simulated 
annealing, help in supervised and unsupervised tasks 
\citep{geras2014scheduled, chandra2014adaptive, rennie2014annealed}.
We propose an alternative that avoids a schedule and rather
uses random noise levels such that the model cannot adapt to slowly changing 
noise distribution.

A similar approach of sampling the noise level was used in \cite{geras2014scheduled} in the context
of unsupervised learning using an autoencoder (on input not features).
However, they show that the approach is not very useful in their case.

Finally, work by \cite{graham2015efficient} is related to random noise as 
both their submatrix multiplications and random noise level $\rho$ are inducing
an independence between neurons of a single layer. They found that the 
independence is not to damageable if they use enough different submatrix patterns.

%% file: text/conclusion.tex
We have presented and justified a novel interpretation of dropout as
prior-knowledge free data augmentation. We described a new procedure to 
generate samples by back projecting the dropout noise into the input space.
Our results suggest neural networks can be trained without dropout on such
noisy samples and still yield good results. Nonetheless, experiments should be
performed on larger networks in order to determine whether this observation is
just a particular property of relatively small networks. Furthermore, trained 
networks should be analyzed to determine if co-adaptation is still avoided when 
using per-layer noise back-projection on deep neural networks. 

Presenting only random dropout, the list of possible substitute to dropout in 
this work is far from exhaustive. As described in section \ref{sec:improve}, 
important knobs to modify induced data augmentation by noise are model's
features and the noise scheme applied on them. Using semi-supervised cost can
influence the implicit transformations by forcing the network to learn more
general features. A network could also be trained on $x^*$ samples generated
from another network, similarly to generative adverserial networks
\citep{goodfellow2014generative}.

%% file: text/appendix.tex
\subsection{Proof of $x^*$ unlikeliness}
\label{subsec:x_star_proof}
We can show, with a proof by contradiction, that it's unlikely to find a single 
$x^* = x^{(1)^*} = x^{(2)^*} = \dots = x^{(n)^*}$ that minimizes well $L$. 

By the associative property of function composition, we can rewrite equation \ref{eq:f_x_star}

\begin{equation}
    f^{(i)}(x^*)  = \left(a ^{(i)} \circ h^{(i)}\right)\left(f^{(i-1)}\left(x^*\right)\right) \label{eq:f_x_star_recur}
\end{equation}

Suppose there exist an $x^*$ such that 

\begin{align}
    \left(a^{(i)}   \circ h^{(i)}  \right)\left(f^{(i-1)}\left(x^*\right)\right) &= \left(\tilde{a}^{(i)}     \circ h^{(i)}\right)\left(\tilde{f}^{(i-1)}(x)\right) \label{eq:equal-i}\\
    \left(a^{(i-1)} \circ h^{(i-1)}\right)\left(f^{(i-2)}\left(x^*\right)\right) &= \left(\tilde{a}^{(i-1)} \circ h^{(i-1)}\right)\left(\tilde{f}^{(i-2)}(x)\right) \label{eq:equal-i-1}
\end{align}

Based on \ref{eq:f_x_star_recur} and \ref{eq:equal-i-1}, we have that $f^{(i-1)}(x^*) = \tilde{f}^{(i-1)}(x)$. The proof is concluded by replacing the latter in \ref{eq:equal-i} and
then expanding the composed functions.

\begin{align}
    \left(a^{(i)}   \circ h^{(i)}  \right)\left(f^{(i-1)}\left(x^*\right)\right) &= \left(\tilde{a}^{(i)}     \circ h^{(i)}\right)\left(f^{(i-1)}(x^*)\right) \nonumber\\
           \mbox{rect}\left(h^{(i)}\left(f^{(i-1)}\left(x^*\right)\right)\right) &= \vec{m}^{(i)}\odot\mbox{rect}\left(h^{(i)}\left(f^{(i-1)}\left(x^*\right)\right)\right) \label{eq:conclusion}
\end{align}

Equation \ref{eq:conclusion} can only be true if $\vec{m}^{(i)}$ does not apply 
any modification to $\mbox{rect}\left(h^{(i)}\left(f^{(i-1)}\left(x^*\right)\right)\right)$, 
that means $\vec{m}^{(i)}_j=1$ when 
$\mbox{rect}_j\left(h^{(i)}\left(f^{(i-1)}\left(x^*\right)\right)\right) > 0$. 
It happens with a probability $p_{(i)}^{d_{(i)}s_{(i)}}$ 
where $p_{(i)}$ is the Bernoulli success probability, $d_{(i)}$ is the number 
of hidden units and $s_{(i)}$ is the mean sparsity level, i.e. mean percentage of 
active hidden units, of the $i$-th hidden layer. This probability is very low for 
standard hyper-parameters values. For instance, with $p_{(i)}=0.5$, $d_{(i)}=1000$ 
and $s_{(i)}=0.15$, the probability is as low as $10^{-47}$.